\documentclass[a4paper]{article}

\usepackage{INTERSPEECH2019}
\usepackage{cite,subcaption}
\usepackage{balance}
\usepackage[dvipsnames]{xcolor}
\title{Code-Switching Detection Using ASR-Generated Language Posteriors}
\name{Qinyi Wang, Emre Y\i lmaz, Adem Derinel, Haizhou Li}
\address{
  Dept. of Electrical and Computer Engineering, National University of Singapore, Singapore}
\email{qinyi@u.nus.edu}

\begin{document}

\maketitle
\begin{abstract}
Code-switching (CS) detection refers to the automatic detection of language switches in code-mixed utterances. This task can be achieved by using a CS automatic speech recognition (ASR) system that can handle such language switches. In our previous work, we have investigated the code-switching detection performance of the Frisian-Dutch CS ASR system by using the time alignment of the most likely hypothesis and found that this technique suffers from over-switching due to numerous very short spurious language switches. In this paper, we propose a novel method for CS detection aiming to remedy this shortcoming by using the language posteriors which are the sum of the frame-level posteriors of phones belonging to the same language. The CS ASR-generated language posteriors contain more complete language-specific information on frame level compared to the time alignment of the ASR output. Hence, it is expected to yield more accurate and robust CS detection. The CS detection experiments demonstrate that the proposed language posterior-based approach provides higher detection accuracy than the baseline system in terms of equal error rate. Moreover, a detailed CS detection error analysis reveals that using language posteriors reduces the false alarms and results in more robust CS detection.
\end{abstract}


\section{Introduction}
Code-switching (CS), the alternating use of two or more languages in a single conversation, is a common phenomenon in multilingual communities. There is increasing research interest in developing CS automatic speech recognition (ASR) systems~\cite{li2012,adel2013,adel2014,zeng2017,hamed2017,westhuizen2017,stemmer2001,lyu2006,vu2012,modipa2013,yilmaz2016_2,weiner2012,lyu2013,yeong2014} as most of the off-the-shelf systems are monolingual and cannot handle code-switched speech. Our previous research has focused on developing an all-in-one CS ASR system using a Frisian-Dutch bilingual acoustic and language model that allows language switches~\cite{yilmaz2016_2,yilmaz2016_4}.

Performing CS detection on a code-switched speech can automatically determine the points of code-switching and language identities of words in the code-switched utterances which can eventually be used for speech recognition tasks, such as spoken term detection and improving the CS ASR performance. Among two main types of code-switching, namely inter-sentential (which occurs between sentences) and intra-sentential (which occurs within a single sentence) \cite{myers1989}, the detection of the latter is more challenging due to shorter duration between the CS points.

Despite considerable research effort on language recognition (LR) and diarization, there is little previous work available on CS detection~\cite{volk2014,wu2015,yilmaz2016_4,yilmaz2017_1,amazouz2017}. It is worth mentioning that the CS detection task is more challenging than the standard LR experimental setting due to: (1) considerably shorter monolingual segment duration (i.e., intra-sentential CS durations can be as short as a few seconds) and (2) uncertainty over the language boundaries.

One previous work on detecting language switches uses multiple monolingual ASR systems in parallel and assigns language identities to words with the language of the system with highest likelihood score \cite{amazouz2017}. In our earlier work, we have described a CS detection technique which uses a Frisian-Dutch CS ASR system to recognize the most likely transcription of each utterance and detect language switches based on the time alignment of the ASR output~\cite{yilmaz2016_4}. 

In our latest work, we have investigated this code-switching detection performance using data-augmented CS ASR systems and observed that this technique suffers from over-switching due to numerous very short spurious language switches~\cite{yilmaz2018_3}. To cope with shortcoming, this paper introduces a new method for code-switching detection which uses frame-level language posteriors which are created using a CS ASR system. This technique extracts a posterior probability for each language by summing the phone posteriors belonging to the same language. A frame-level decision is then made by choosing the language with the highest language posterior.

We present CS detection results on the FAME! corpus~\cite{yilmaz2016} using various detection systems including (1) the baseline system relying on the time alignment of the ASR output, (2) an intermediate system which makes a decision based on the maximum phone posteriors, and (3) the proposed language posterior-based technique. After reporting the detection performance, an analysis of the hypothesized CS by each technique is provided by reporting the total number of hypothesized language switches and duration distribution of the monolingual segments to provide further insight into the quality of the hypothesized language switches. The proposed technique has not only given lower equal error rates but also hypothesized language switches that most closely resembles the human annotations compared to the baseline CS detection technique.

The rest of the paper is organized as follows. Section~\ref{sec:CSdet} details the baseline and proposed CS detection techniques and Section~\ref{sec:expset} summarizes the experimental setup and implementation details. Section~\ref{sec:res} presents and discusses the CS detection results and analyses before the conclusion given in Section~\ref{sec:conc}.

\section{Frisian-Dutch Radio Broadcast Database}
\label{sec:database}
West Frisian is one of the three Frisian languages (together with East and North Frisian spoken in Germany) and it has approximately half a million speakers mostly living in the province Frysl\^{a}n located in the northwest of the Netherlands. The native speakers of West Frisian (Frisian henceforth) are mostly bilingual and often code-switch in daily conversations due to the extensive influence of the Dutch language \cite{popkema2013}.

The bilingual FAME! speech database has been collected in the scope of the \textit{Frisian Audio Mining Enterprise} project and contains radio broadcasts in Frisian and Dutch from the archive of the regional public broadcaster Omrop Frysl\^{a}n (Frisian Broadcast Organization). This bilingual data contains Frisian-only and Dutch-only utterances as well as mixed utterances with inter-sentential, intra-sentential and intra-word CS. These recordings include language switching cases and speaker diversity, and have a large time span (1966--2015). The longitudinal and bilingual nature of the material enables to perform research into language variation in Frisian over years, formal versus informal speech, language change across the life-span, dialectology, code-switching trends, speaker tracking and diarization over a large time period. For further details, we refer the reader to~\cite{yilmaz2016}.

\section{CS detection techniques}
\label{sec:CSdet}

The code-switching detection task involves detection of language boundaries (switching points) and identification of the language identity of preceding and following subsegments in a code-switched utterance. The following subsections details the baseline CS detection technique which uses the time alignment of the most likely ASR hypothesis and the techniques relying on phone and language posteriors.

\subsection{Baseline approach: time alignment of CS ASR output}

One way of obtaining frame-level language labels hypothesized by a CS ASR is to align the most likely hypothesis and assign each frame language labels using language tags appended to the words. CS ASR employs a bilingual acoustic model that captures the phonetic characteristics of both languages and a bilingual language model (LM) which can assign probabilities to code-mixed word sequences as well as monolingual word sequences from both languages. The current system uses data-augmented models described in \cite{yilmaz2018_2}. The acoustic model is trained on automatically transcribed data from the same archive and a large amount of monolingual data from the high-resourced language (Dutch) together with the manually transcribed data form the FAME! corpus. Moreover, we have created CS text, which is almost nonexistent, in multiple ways providing perplexity reductions on the development and test set transcriptions. The data-augmented models have been shown to provide better CS detection in terms of equal error rate, but have the tendency to hypothesize much more language switches compared to the human annotations~\cite{yilmaz2018_3}.

\subsection{Proposed approach: language posteriors}

Rather than relying on the most likely hypothesis, code-switching detection can also be achieved at frame level using the phone posteriors. Each phone having a language tag, the phones tagged with the same language are further summed to obtain a posterior probability for each language. The CS ASR-generated language posteriors contain more complete language-specific information at frame level than the time alignment of the most likely hypothesis. Therefore, using language posteriors for CS detection is expected to yield more accurate and robust CS detection. The silences are considered as a third class as they do not belong to any language. To reduce the confusion of acoustically similar phones in both languages, a phone LM is incorporated during the phone posterior extraction. 

\begin{figure}[!t]
\centering
\begin{subfigure}{.5\textwidth}
  \centering
  \includegraphics[trim=1cm 0cm 0cm 0cm,width=0.9\linewidth]{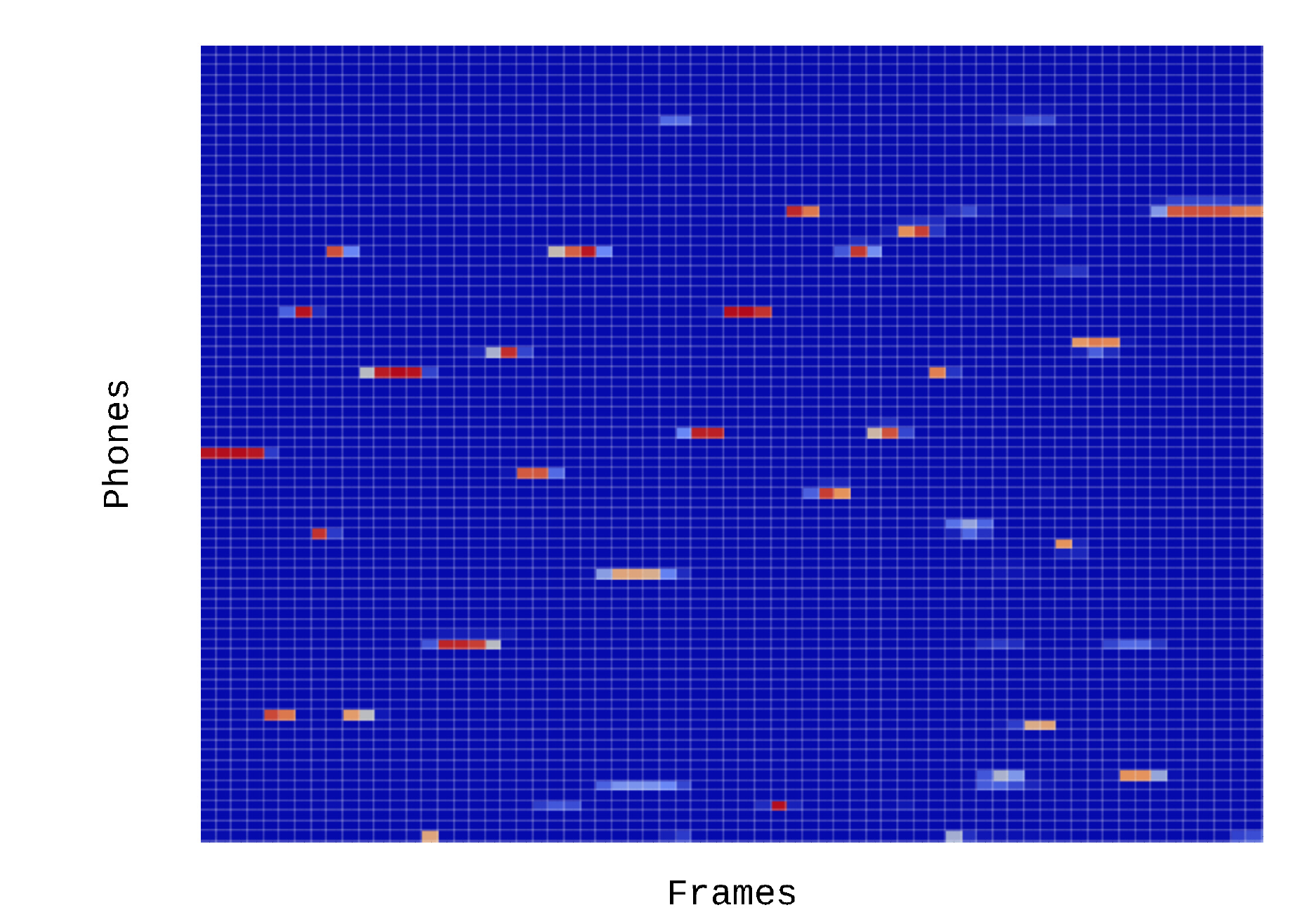}
  \caption{Phone posteriors}
  \label{fig:sub1}
\end{subfigure}
\begin{subfigure}{.5\textwidth}
  \centering
  \includegraphics[trim=1cm 0cm 0cm 0cm,width=0.9\linewidth]{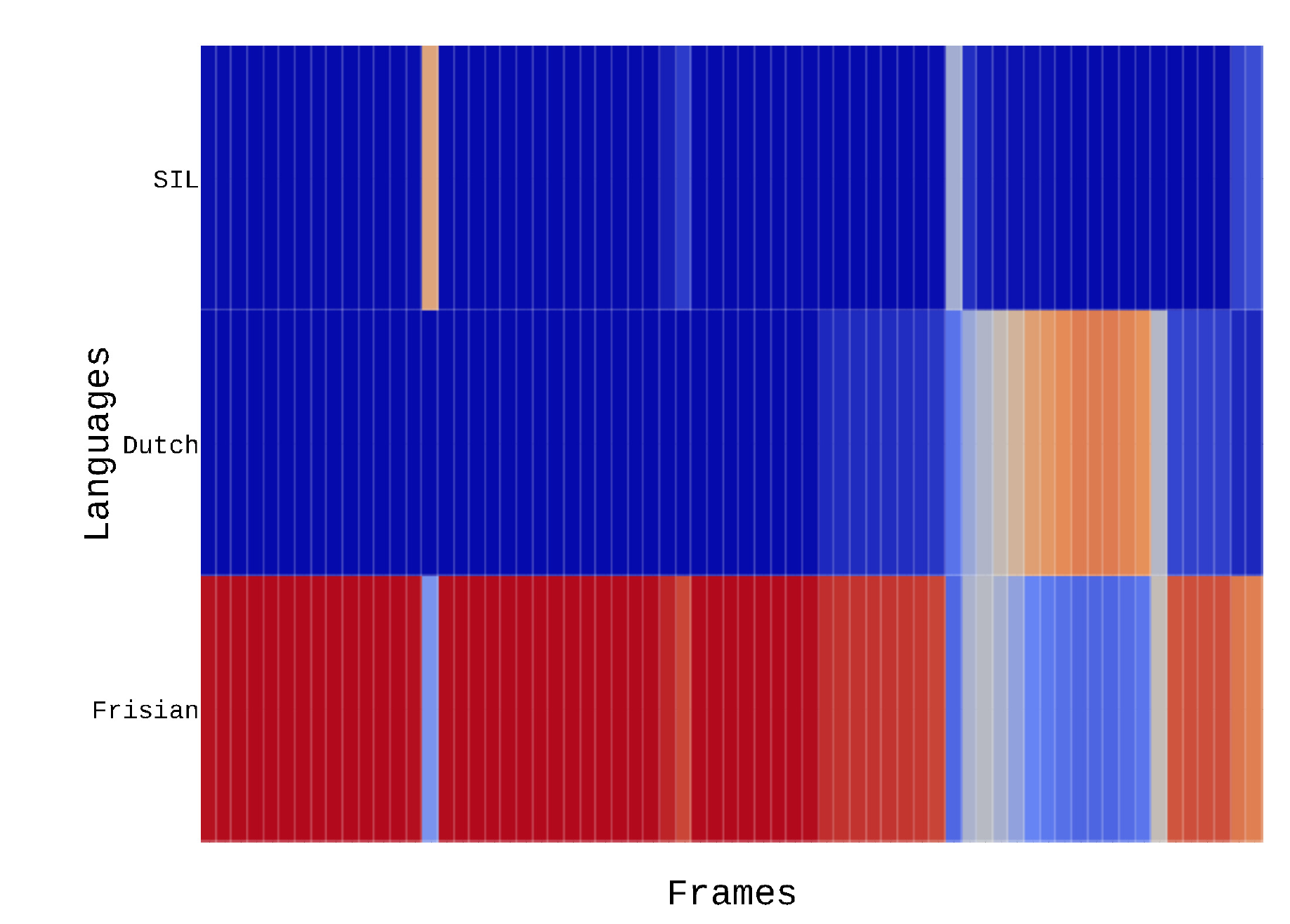}
  \caption{Language posteriors}
  \label{fig:sub3}
\end{subfigure}
\caption{Phone posteriors of a code-switched utterance and the corresponding language posteriors obtained by summing all same-language phone posteriors}
\vspace{-0.5cm}
\label{fig:ppg}
\end{figure}

The language decision per frame is made in two ways based on: (1) the language tag of the phone with the highest posterior and (2) the highest language posterior. Figure \ref{fig:ppg} illustrates an example of phone posteriors for a code-switched utterance with the corresponding language decision based on the maximum phone and language posteriors. The CS detection based on the maximum phone posterior is more susceptible to uncertainty between phones than the maximum language posterior as it can be seen from the example. Summing the posterior probabilities assigned to all phones of a language yields a more reliable evidence for the language identity resulting in a more robust CS detection.

%

\begin{figure}[t]
  \centering
  \includegraphics[trim=0.5cm 0.75cm 1cm 2cm, width=1\linewidth]{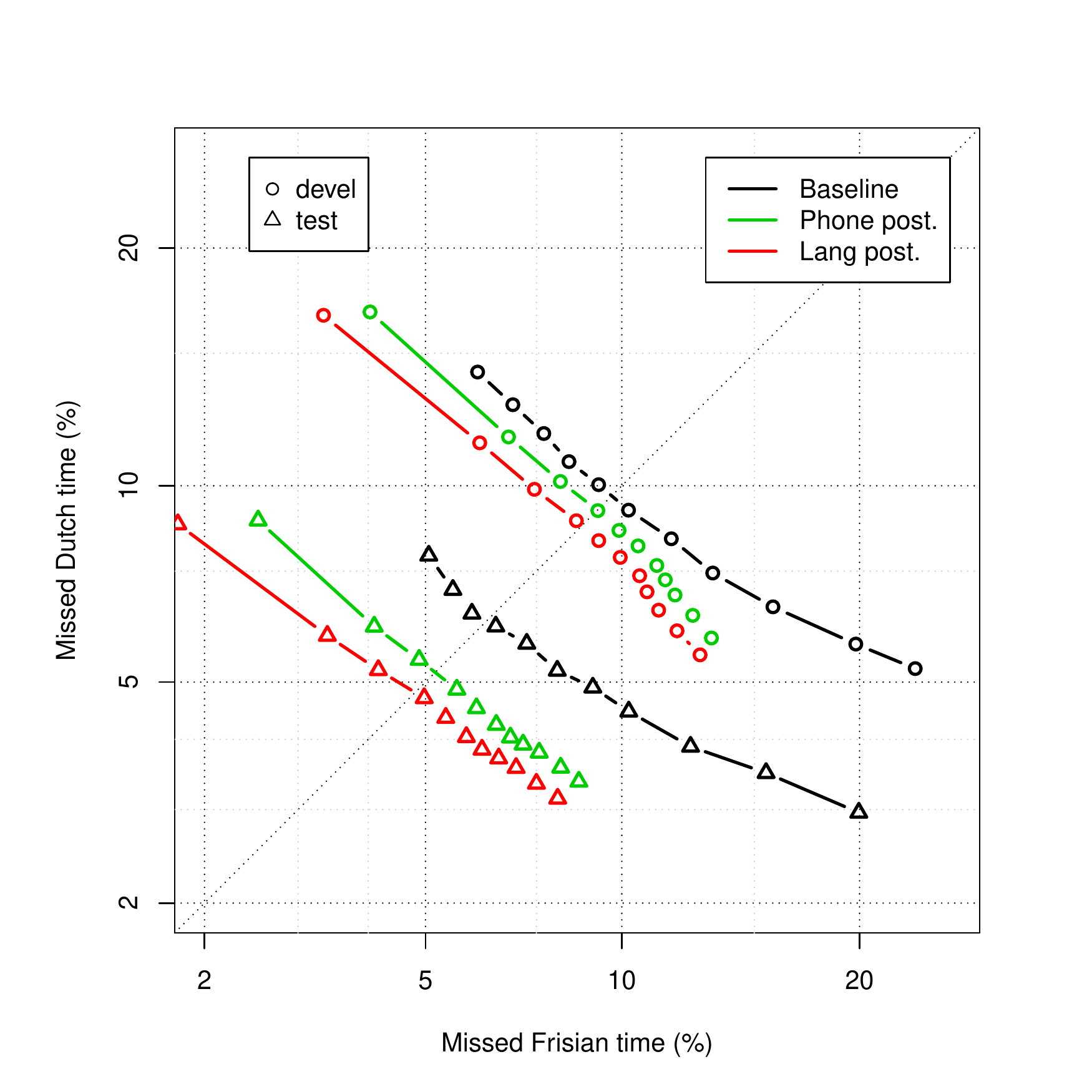}
  \caption{Performance of code-switching detection systems implemented with different methods on FAME! development and test data}

  \label{fig:det}
\end{figure}

\section{Experimental Setup}
\label{sec:expset}
\subsection{Speech and text data}

The training data of the FAME! speech corpus comprises 8.5 hours and 3 hours of speech from Frisian and Dutch speakers respectively. The development and test sets consist of 1 hour of speech from Frisian speakers and 20 minutes of speech from Dutch speakers each. All speech data has a sampling frequency of 16 kHz. The amount of automatically annotated speech data extracted from the target broadcast archive is 125.5 hours. 

Monolingual Dutch speech data comprises the complete Dutch and Flemish (language variety spoken in Belgium) components of the Spoken Dutch Corpus (CGN)~\cite{cgn} that contains diverse speech material including conversations, interviews, lectures, debates, read speech and broadcast news. This corpus contains 442.5 and 307.5 hours of Dutch and Flemish data respectively.

The bilingual text corpus used for LM training contains 107.3M words in total (monolingual Frisian text with 37M words, monolingual Dutch text with 8.8M Dutch words and automatically generated CS text with 61.5M words). Almost all Frisian text is extracted from monolingual resources such as Frisian novels, news articles, Wikipedia articles. The Dutch text is extracted from the transcriptions of the CGN speech corpus which has been found to be very effective for LM training compared to other text extracted from written sources. The transcriptions of the FAME! training data is the only source of CS text and contains 140k words. The remaining CS text is automatically generated as described in \cite{yilmaz2018_2}.

\subsection{Implementation details}

The CS ASR system used in these experiments is trained using the Kaldi ASR toolkit~\cite{kaldi}. We train a conventional context dependent Gaussian mixture model-hidden Markov model (GMM-HMM) system with 40k Gaussians using 39 dimensional mel-frequency cepstral coefficient (MFCC) features including the deltas and delta-deltas to obtain the alignments for training a lattice-free maximum mutual information (LF-MMI)~\cite{povey2016} TDNN-LSTM~\cite{peddinti2017} AM (1 standard, 6 time-delay and 3 LSTM layers). We use 40-dimensional MFCC as features combined with i-vectors for speaker adaptation~\cite{saon2013}. The LM used for the baseline CS detection system is a standard bilingual 3-gram with interpolated Kneser-Ney smoothing. Further details are provided in \cite{yilmaz2018_2}. We compute phone posteriors from the denominator graph (created using a phone LM estimated from the phone alignments of the training data) of the chain model and map them to phones using the existing implementation in Kaldi (\textit{nnet3-chain-compute-post}). The output obtained for each frame is $l_{1}$-normalized and the resulting normalized vectors are used for CS detection.

The bilingual lexicon contains 110k Frisian and Dutch words. The number of entries in the lexicon is approximately 160k due to the words with multiple phonetic transcriptions. In this version of the CS ASR system, we have updated the spelling of certain Frisian words in the text corpora, pronunciation lexicon and transcriptions according to the latest spelling rules proposed in 2016 by Fryske Akademy, which is the main difference compared to the previous system in \cite{yilmaz2018_2}. The phonetic transcriptions of the words that do not appear in the initial lexicons are learned by applying grapheme-to-phoneme (G2P) bootstrapping~\cite{davel2003,maskey2004}. The lexicon learning is carried out only for the words that appear in the training data using the G2P model learned on the corresponding language. We use the Phonetisaurus G2P system~\cite{novak2015} for creating phonetic transcriptions. This CS ASR system provided a word error rate of 24.9\% and 23.0\% on the development and test set of the FAME! speech corpus, respectively.

\begin{table}
    \centering
    \caption{EER (\%) provided by different CS detection systems on the development and test data}
    \vspace{-0.15cm}
    \begin{tabular}{| l | c | c |} 
    \hline
    CS Detection System & Development & Test  \\
    \hline\hline
    Baseline~\cite{yilmaz2018_3} & 9.7 & 6.3 \\
    \hline
    Phone posterior & 9.2 & 5.2 \\
    \hline
    Language posterior & 8.7 & 4.8 \\
    \hline
    \end{tabular}
    \label{tab:eer}
\end{table}

\subsection{CS detection experiments}

For the baseline CS detection system, we trained a monolingual Frisian and Dutch LM, and interpolated between them with varying weights. This effectively varies the prior for the detected language. The most likely hypothesis is obtained for each utterance using each interpolated LM and its time alignment on phone level is stored in a .ctm format. By comparing these alignments with the ground truth word-level alignments (obtained by applying forced alignment using the recognizer), a duration-based CS detection accuracy metric has been calculated. The missed Frisian (Dutch) time is calculated as the ratio of total duration of frames with Frisian (Dutch) tag in the reference alignment which is aligned to frames without Frisian (Dutch) tag to the total number of frames with Frisian (Dutch) tag in the reference alignment. 

The same procedure is followed for the approaches using phone and language posteriors, except the use of interpolated phone LMs with varying weights for manipulating the language priors. After extracting the phone posteriors for each utterance using different phone LMs, a language label is assigned to each frame based on the maximum phone or language posteriors. These frame-level language labels are stacked in a vector for each utterance and later converted to a .ctm file which marks the start times and durations of each monolingual segment. The same duration-based metric is used to evaluate the CS detection accuracy.

The CS detection accuracy is evaluated by reporting the equal error rates (EER) calculated based on the detection error tradeoff (DET) graph \cite{martin1997} plotted for visualizing the CS detection performance. In previous work, we have observed that this duration-based CS detection metric penalizes very short erroneous language switches less compared to incorrect language tags assigned over longer segments which gives a better indication of the general language recognition capability of the corresponding ASR system. Therefore, we further analyze the number of hypothesized language switches and duration distribution of the monolingual segments to gain insight about the CS detection behavior of each system.

\section{Results and Discussion}
\label{sec:res}

The EERs and DET curves provided by the three CS detection systems are given in Table \ref{tab:eer} and Figure \ref{fig:det}, respectively. The baseline CS detection system has an EER of 9.7\% on the development and 6.3\% on the test set. CS detection based on the language tag of phone with the maximum phone posterior has a reduced EER of 9.2\% on the development and 5.2\% on the test data. The lowest EERs are given by the system using the language posteriors. The consistent improvements on the development and test sets indicate the improved overall CS performance over the baseline and phone posterior-based system.

\begin{figure}[t]
  \centering
  \includegraphics[trim=0cm 0cm 1cm 0.5cm, width=1.05\linewidth]{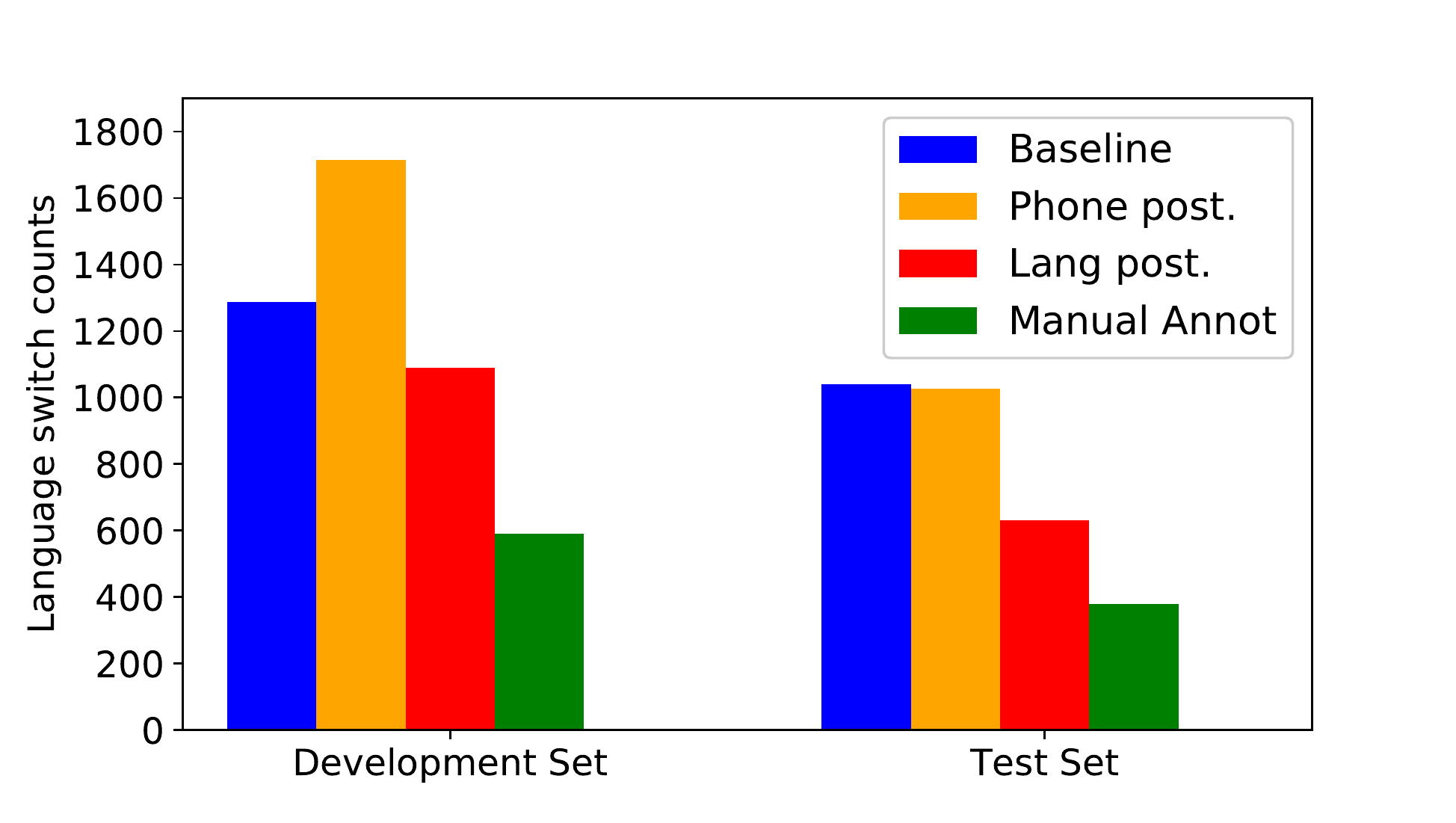}
  \caption{Hypothesized language switch counts}
  \vspace{-0.2cm}
  \label{fig:switch}
\end{figure}

\begin{figure}[t]
  \centering
  \includegraphics[trim=1cm 0.5cm 1cm 0.5cm, width=1.05\linewidth]{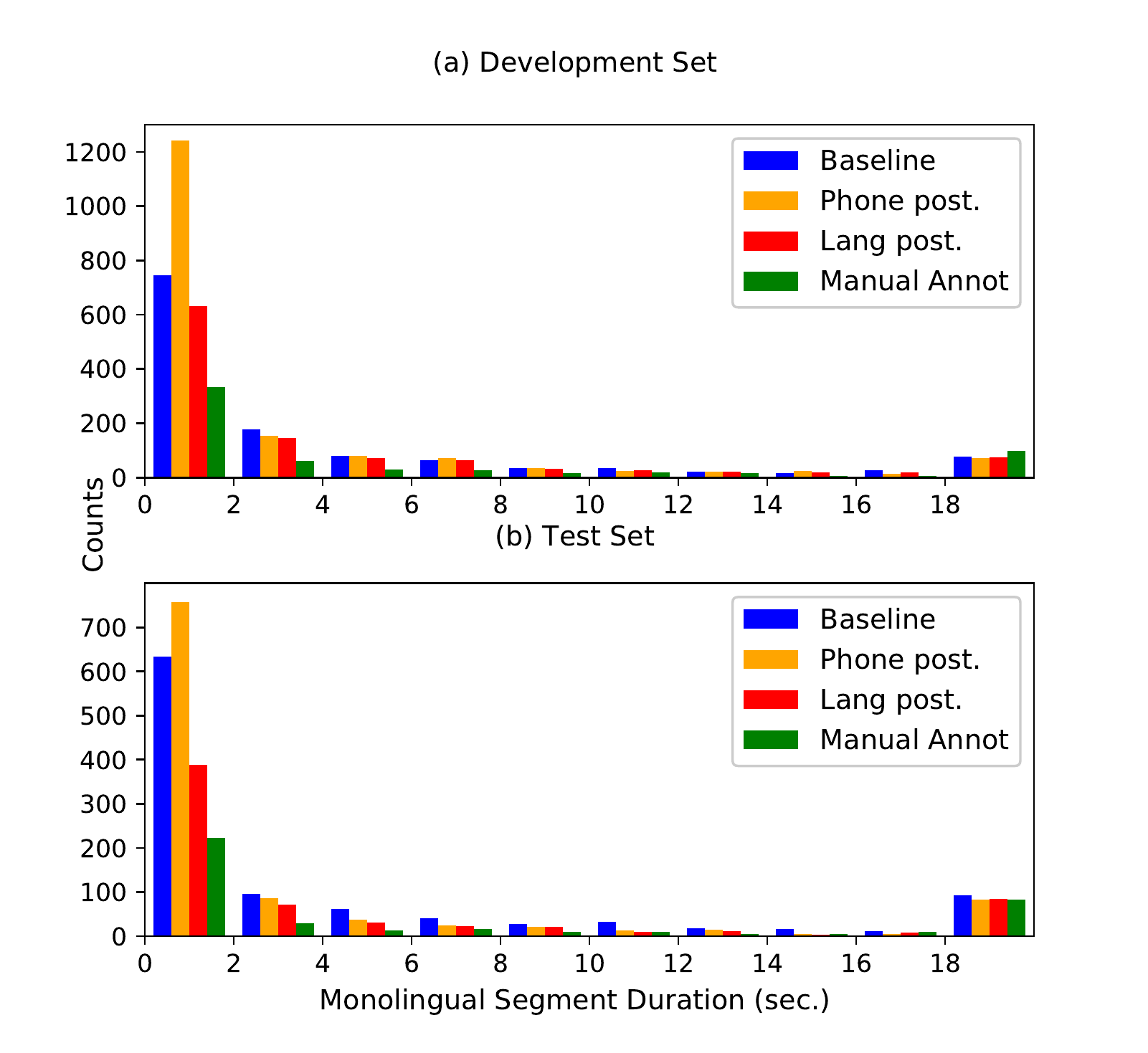}
  \caption{Duration distribution of monolingual segments}
  \vspace{-0.2cm}
  \label{fig:duration}
\end{figure}

We further compare the number of language switches hypothesized by each CS detection system with the manually annotated switches in Figure \ref{fig:switch}. Additionally, the histograms of the duration distributions of the monolingual speech segments are shown in Figure \ref{fig:duration}. These plots reveal that all CS detection systems tend to overestimate the number of language switches in the code-switched speech. These false alarms are mainly due to a large number of monolingual segments that are shorter than 2 seconds as shown in Figure \ref{fig:duration}. Using language posteriors helps reducing the amount of false alarms on both sets. 

In general, the proposed language posterior-based CS detection system provides a higher overall CS detection accuracy than the baseline technique using the time alignment of the mostly likely hypothesis. Although all systems still suffer from false alarms (by incorrectly hypothesizing very short-duration language switches), using a language posterior-based CS detection system alleviates this problem. This indicates that the proposed CS detection system is more robust compared with other systems. Further investigation needs to be done to reduce these false alarms which remains as a future work.

\section{Conclusion}
\label{sec:conc}
This paper addresses code-switching (CS) detection problem and introduces a new method for CS detection by using frame-level language posteriors produced by a CS ASR system. The language posteriors are obtained by summing the posteriors of the same-language phones extracted using the bilingual acoustic model in conjunction with a phone language model. The performance of this CS detection system is compared with a baseline system that uses the time alignment of the most likely hypothesis. The CS detection experiments indicate that the proposed CS detection system provides the lowest EERs on development and test set of the FAME! corpus. We also demonstrate that using language posterior for CS detection yields more robust detection with a considerably reduced number of false alarms due to incorrectly hypothesized very short language switches. Future work directions include investigating smoothing techniques to address the overestimation problem and exploiting the CS detection results to improve the CS ASR performance.

\section{Acknowledgements}
This research is supported by National Research Foundation through the AI Singapore Programme, the AI Speech Lab: Automatic Speech Recognition for Public Service Project AISG-100E-2018-006.

\bibliographystyle{IEEEtran}
\balance
\bibliography{qinyi-refs}


\end{document}